\documentclass[a4paper, 10pt, conference]{ieeeconf}      % Use this line for a4
\usepackage{times}
\usepackage{epsfig}
\usepackage{graphicx}
\usepackage{amsmath}
\usepackage{amssymb}
\usepackage{booktabs}
\usepackage{lipsum}
\usepackage{multirow}
\usepackage{subfloat}
\usepackage{subcaption}
\usepackage{xcolor}

\IEEEoverridecommandlockouts                              % This command is only
\title{\LARGE \bf \vspace{.2cm}
Latent Embedding Clustering for \\ Occlusion Robust Head Pose Estimation
}

\author{\parbox{16cm}{\centering
    {\large José Celestino, Manuel Marques and Jacinto C. Nascimento}\\
    {\normalsize
    Institute for Systems and Robotics, Instituto Superior Técnico, Lisboa, Portugal\\}}
}

\usepackage{fancyhdr}
\begin{document}

%\ifFGfinal
%\thispagestyle{empty}
%\pagestyle{empty}
%\else
%\author{Anonymous FG2024 submission\\ Paper ID 181 \\}
%\pagestyle{plain}
%\fi
\maketitle

\thispagestyle{fancy}

%%%%%%%%%%%%%%%%%%%%%%%%%%%%%%%%%%%%%%%%%%%%%%%%%%%%%%%%%%%%%%%%%%%%%%%%%%%%%%%%
\begin{abstract}

    Head pose estimation has become a crucial area of research in computer vision given its usefulness in a wide range of applications, including robotics, surveillance, or driver attention monitoring. One of the most difficult challenges in this field is managing head occlusions that frequently take place in real-world scenarios.
    In this paper, we propose a novel and efficient framework that is robust in real world head occlusion scenarios. In particular, we propose an unsupervised latent embedding clustering with regression and classification components for each pose angle. The model optimizes latent feature representations for occluded and non-occluded images through a clustering term while improving fine-grained angle predictions. %objective function, while improving fine-grained angle predictions.
    Experimental evaluation on in-the-wild head pose benchmark datasets reveal competitive performance in comparison 
    to state-of-the-art methodologies
    %to a occlusion focused method which constitutes our upper-bound target performance, 
    with the advantage of having a significant data reduction. We observe a substantial improvement in occluded head pose estimation. Also, an ablation study is conducted to ascertain the impact of the clustering term within our proposed framework. %objective within our training framework. 

\end{abstract}

%%%%%%%%%%%%%%%%%%%%%%%%%%%%%%%%%%%%%%%%%%%%%%%%%%%%%%%%%%%%%%%%%%%%%%%%%%%%%%%%
\section{INTRODUCTION}
\label{section:Introduction}

\begin{figure}[ht]
  \centering
  \begin{minipage}[b]{0.45\textwidth}
    \centering
    \includegraphics[width=1\textwidth]{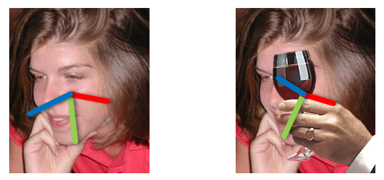}
    \subcaption{}
    \label{subfig:intro1}
  \end{minipage}
  
  \vspace{0.1cm}
  
  \begin{minipage}[b]{0.45\textwidth}
    \centering
    \includegraphics[width=1\textwidth]{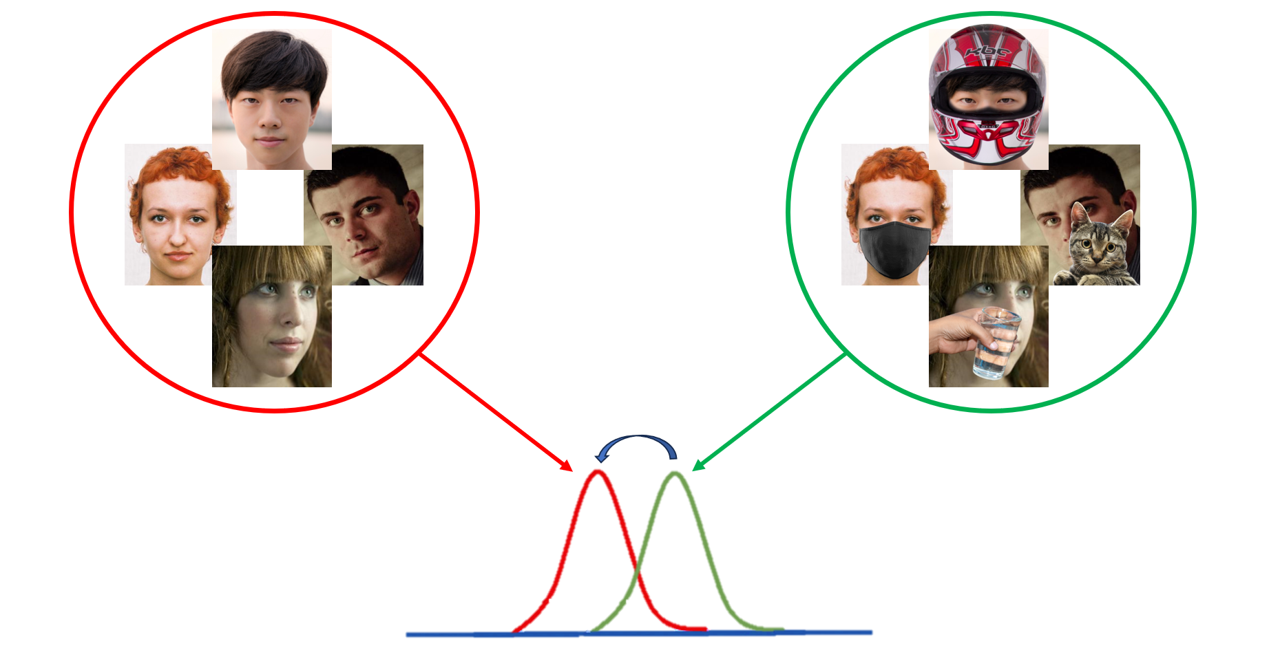}
    \subcaption{}
    \label{subfig:intro2}
  \end{minipage}
  \caption{(a) Occlusions strongly affect state-of-the-art methods for head pose estimation. 
  (b) We propose to improve the occluded HPE estimation by introducing a novel framework containing a multi-loss Euler framework with an unsupervised clustering of the latent space. This is achieved through a minimization of the difference between probability distributions for occluded and non-occluded images.}
  \label{fig:twosubfigures}
  %\vspace{-4mm}
\end{figure}

Head pose estimation (HPE) can be roughly defined as the prediction of the relative orientation (and position) of the human’s head with respect to the camera. HPE became a relevant topic in computer vision, being crucial in providing crucial information for an ever-growing range of applications, including human-computer/robot interaction~\cite{feedbot}, surveillance systems~\cite{surveillance}, driver attention monitoring~\cite{driver_app,driver_1, pandora}, virtual/augmented reality~\cite{augmented_reality}, health care~\cite{healthcare} and marketing~\cite{marketing}. Many of the applications above suffer from a substantial in-the-wild setback that has scarcely been investigated in detail: {\em the presence of occlusions}. Caused by external objects, facial accessories, or even body parts, they are often inevitable and introduce significant difficulties in capturing reliable facial features leading to inaccurate head pose estimates, ({\it  e.g.} Fig.~\ref{subfig:intro1}). Consequently, existing methods often struggle to handle occlusions effectively, which results in unreliable performance in an unconstrained real-world environment. 

\begin{figure}[ht]
  \centering
  \begin{minipage}[b]{0.45\textwidth}
    \centering
    \includegraphics[width=0.9\textwidth]{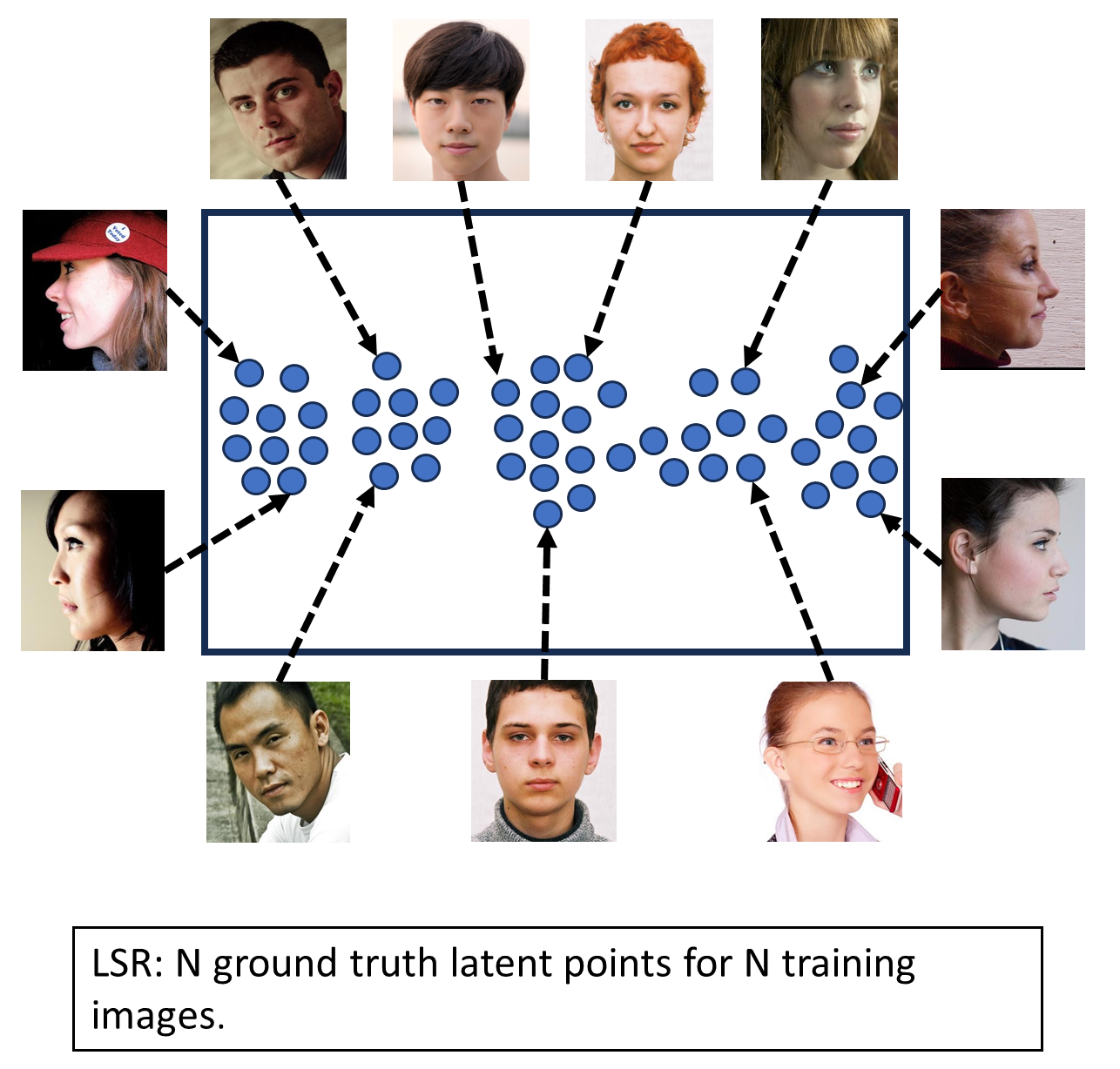}
    \subcaption{}
    \label{subfig:lsr_intro}
  \end{minipage}
  
  \vspace{0.1cm}
  
  \begin{minipage}[b]{0.45\textwidth}
    \centering
    \includegraphics[width=0.9\textwidth]{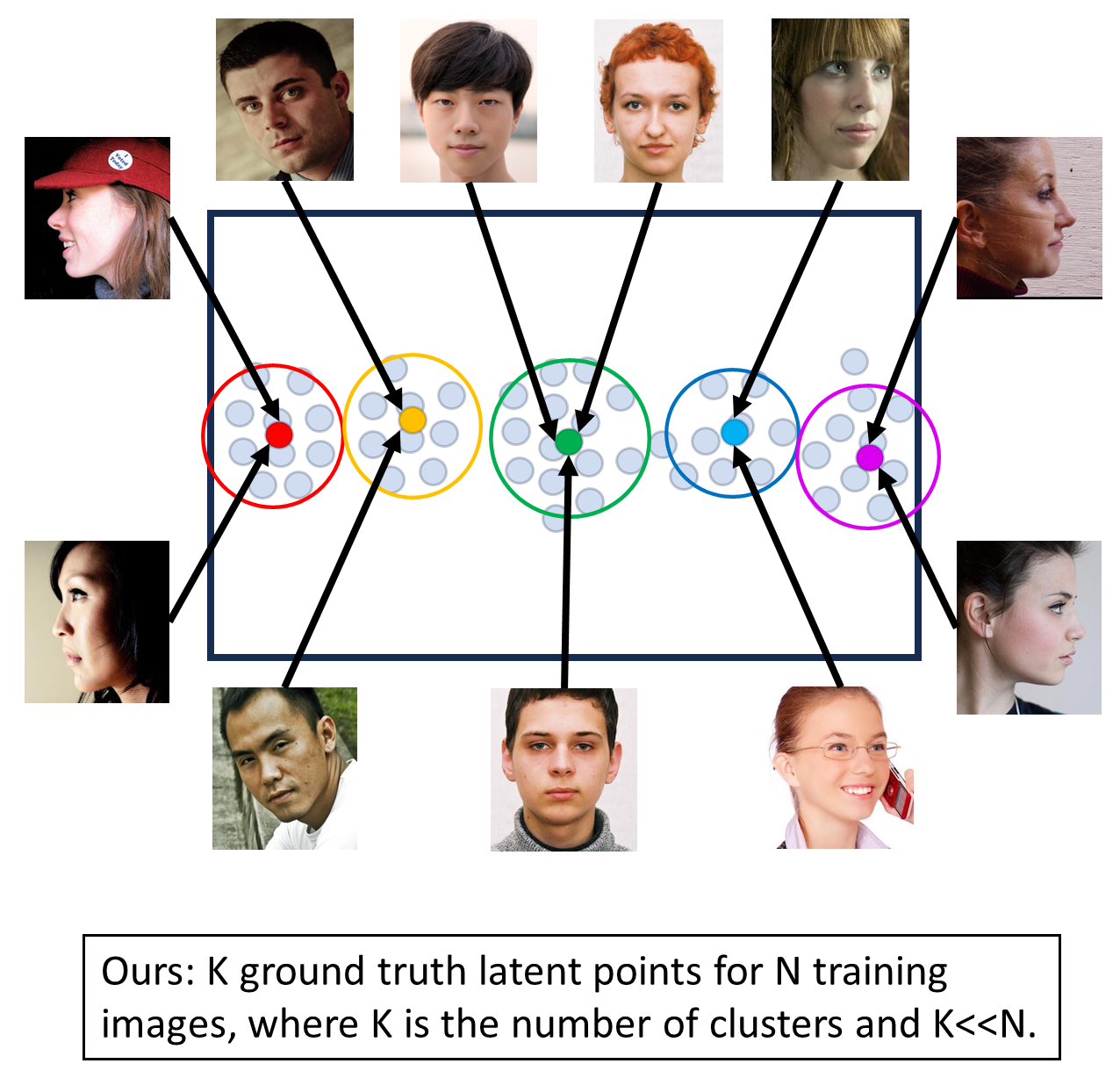}
    \subcaption{}
    \label{subfig:lec_intro}
  \end{minipage}
  \caption{Comparison between~\cite{latent} (a) and our proposed framework (b). While in~\cite{latent}, $N$ ground truth latent points are needed for training with $N$ images, our method only requires $K<<N$ ground truth points to train with the same $N$ images. In our proposal, $K$ are cluster centers points in the latent space.}
  \label{fig:lsr vs lec}
  %\vspace{-4mm}
\end{figure}

%\begin{figure}[t]
%  \centering
%  \begin{minipage}[b]{0.49\textwidth}
%    \centering
%    \includegraphics[width=1\textwidth]{Images/LSR VS LEC NEW FULL cut.png}
%  \end{minipage}
%  \caption{Comparison between~\cite{latent} (left) and our proposed framework (right). While in~\cite{latent}, $N$ ground truth latent points are needed for training $N$ images, our method only requires  $K<<N$ ground truth points, for training the same $N$ images. In our proposal $K$ are cluster centers points in the latent space.}
%  \label{fig:lsr vs lec}
%\vspace{-5mm}
%\end{figure}

\par Deep learning has become a crucial and an unavoidable tool to address this task \cite{6drepnet,whenet,FSA_NET,dad3d}. One of the most recent trends in HPE is the use of {\em latent space regression} (LSR), {\em e.g.}~\cite{latent} that has shown to be effective in improving the robustness and generalization to occluded and non-occluded images. Basically, this methodology applies regression for latent embeddings so that the model learns a similar latent representation for occluded and non-occluded samples of the same pose. 
\par However, we can point out two main limitations of this approach: {\bf(1)} it has a high computational cost since 
it is a fully supervised learning approach requiring the non-occluded ground truth latent space sample for each occluded replica, and {\bf(2)} for data augmentation purposes it always requires occluded/non-occluded image pairs to extract the correspondent non-occluded ground truth labels. We term this as {\em constrained data augmentation}.

% CORTEI AS 3 LINHAS SEGUINTES - JÁ ESTAVA SE ESTAVA A FALAR MUITO DE LSR

%The latent space is a low multi-dimensional space that encodes a meaningful representation of the input or external data ({\em i.e.}, images). 

%In this space, samples that are semantically similar have vectors positioned closer to each other in the latent space. 
%Through regression in this space, it is possible to position occluded and non-occluded samples of the same pose closer to each other obtaining a more generalized model. However, this methodology carries a high computational cost since it is a fully supervised learning approach requiring the non-occluded ground truth latent space vector correspondent to each occluded replica for regression in training. 

\par In this paper, we propose a novel methodology to improve occlusion robustness in HPE that does not suffer from the two limitations above. To accomplish this, we draw inspiration from recent work on embedding clustering for deep learning in computer vision~\cite{dec,idec} and combine unsupervised latent embedding clustering with fine-grained Euler angle regression. In this way, we are able to improve feature representation for pose estimation while fine-tuning the latent embedding space via minimization of a clustering loss (Fig.~\ref{subfig:intro2}), without requiring the high cost of having labeled embedding data for each training image.

We can underscore the following advantages of our proposal. First,  
we only require $K<<N$ ground truth latent embedding points for training $N$ images, instead of having one ground truth point per image (see Figure~\ref{fig:lsr vs lec}). Furthermore, our method allows for expansion of the occluded training dataset as it will be detailed in the experimental evaluation (see Sec.\ref{implementation details}).

In summary, our contributions are as follows: 
\vspace{2mm}
\begin{itemize}
    \item Novel methodology with combined optimization of fine-grained Euler angle estimations and unsupervised latent embedding clustering for refinement of the feature embedding space and improvement of estimation robustness against the occlusions,
    \vspace{2mm}
    \item Low computational cost at estimating HPE (see limitation {\bf (1)} above),
    \vspace{2mm}
    \item State-of-the-art results for occluded images in synthetic and natural benchmark datasets,
    \vspace{2mm}
    \item Competitive performance regarding state-of-the art, ({\em e.g.}~\cite{latent}), with a substantial reduction of the ground truth embeddings,
    \vspace{2mm}
    %\item \textcolor{orange}{Ability to expand occluded training dataset without needing occluded/non-occluded image pairs to extract non-occluded ground truth labels.}
    \item Opposing to the {\em constrained data augmentation} (see limitation {\bf(2)} above), our proposal has the ability to augment the occluded training dataset with images that are not occluded replicas of non-occluded images initially used to establish the ground truth labels, and
    \vspace{2mm}
    \item Ablation study regarding the impact of the novel application of unsupervised embedding clustering in the HPE context.
\end{itemize}
\vspace{2mm}
The  paper is organized as follows: Section~\ref{section: Related work} revises related literature. Section ~\ref{section: methodology} describes the proposed methodology to achieve occlusion robustness in HPE. In Section~\ref{section: experimental evaluation}, we compare the performance of the proposed framework with SoTA methodologies in three benchmark datasets: (i) BIWI \cite{biwi}, (ii) AFLW2000 \cite{aflw}, with both occluded and non-occluded images; and (iii) Pandora \cite{pandora}, with real-life occlusions. We also perform ablation studies regarding the novel use of unsupervised embedding clustering. In Section~\ref{section: conclusion}, we present our concluding remarks and potential venues for future improvement.

%%%%%%%%%%%%%%%%%%%%%%%%%%%%%%%%%%%%%%%%%%%%%%%%%%%%%%%%%%%%%%%%%%%%%%%%%%%%%%%%
\section{RELATED WORK}

\label{section: Related work}

In this section, we delve into existing literature of relevant research closely related to our current study. 

%\vspace{1cm}

\subsection{Head Pose Estimation}
Regarding the literature in HPE, the state of the art can be framed in two different classes of strategies: (i) based on facial landmarks detection and model fitting; or (ii) through a deep learning model based on image features.  The existing research is very extensive and diverse for the two classes above.  
\par The model-based class of approaches tries to fit a head mesh to facial landmarks/keypoints collected from the image. Landmarks represent model fixed points that define the contours of specific face regions, ({\it  e.g.} mouth, eyes, nose) in a head mesh. Keypoints, on the other hand, are tracked head feature points that do not have predetermined locations in the image domain. The work in~\cite{hpe_keypoints_landmarks} follows a fusion method that combines landmark and keypoint detection to improve performance when compared to using only one of them alone. Other recent works related to this class of approaches propose fitting landmarks and keypoints to morphable head models, instead of rigid ones. The idea is to improve the facial model fitting to different individuals, since generalization issues may occur in rigid models. The authors of~\cite{ref65} propose a method that requires only four non-coplanar keypoints and includes a 3D face morphing method. The work of ~\cite{wacv2} also aims to estimate the head pose from a small set of head keypoints. Other examples are ~\cite{3DMM}, where pose estimation is approached as a 3D Morphable Model (3DMM) parameter regression problem, and~\cite{dad3d}, which fits a 3D FLAME head mesh and simultaneously learns the head pose, shape and expression.
\par In the wake of advancements in deep learning, new learning-based approaches have emerged for addressing the HPE problem without need of landmarks or keypoints. The authors of~\cite{hopenet} describe the advantages that these methods have over landmark-to-pose methodologies: they do not depend on head meshes, choice of landmark detection or 2D to 3D alignment methods. Several recent works have followed this class of approach, {\em e.g.}~\cite{hopenet, whenet, hpe_3d_nn, FSA_NET, img2pose, quatnet, 6drepnet, lightweight, wacv_1}. The authors of~\cite{FSA_NET} solve HPE as a soft-stage-wise regression problem, inspired by~\cite{ssr_net}. In ~\cite{img2pose}, the authors propose simultaneous face detection and head pose estimation with 6 degrees of freedom. This allows for more robust face detection with low computational cost regarding HPE. The works of~\cite{6drepnet} and~\cite{lightweight} try to address common problems of HPE methods. In~\cite{6drepnet}, the authors try to counter the ambiguity problem of rotation labels by using a rotation matrix representation and a geodesic loss. The method in~\cite{lightweight} tackles the perspective distortion in face images caused by the misalignment of the face with the camera coordinate system. The authors from~\cite{hopenet} propose a deep learning strategy that employs a backbone neural network augmented with three fully-connected layers, each one used to predict a different Euler angle using a multi-loss approach that combines a classification loss with a weighted regression loss for fine-grained predictions. This approach inspired other works, such as~\cite{whenet} which extends the idea to a full $360^\circ$ range of yaws by synthetically extending the training dataset and introducing a new wrapped loss. Instead of regressing Euler angles, the work of ~\cite{quatnet} proposes a quaternion-based multi-regression loss method. This approach incorporates ordinal regression, enabling us to tackle the non-stationary nature of HPE, where variations in pose differ within each angle interval.

\subsection{Occlusions in HPE}

Even though occlusions are one of the roadblocks encountered in HPE, not many works have tackled this problem in depth. The method in~\cite{occluded_hpe1} uses an iterative Lucas-Kanade optical flow tracker to track the displacement of a face feature concerning the center of the head, but still requires the mouth to be free from occlusions. The work in \cite{occluded_hpe2} estimates landmark visibility probabilities to perform occlusion prediction and includes prior occlusion pattern loss to improve performance. The authors of \cite{occluded_hpe3} combine the estimation of facial landmarks, head pose, and facial deformation under  occlusions, but only evaluate yaw estimation. The work in~\cite{latent} includes the generation of synthetically occluded datasets and proposes combining a multi-loss for fine-grained predictions with latent embedding regression. This allows to improve the latent feature representation of synthetically occluded images to that of the original non-occluded images, and therefore, enhances the pose prediction with occlusions. %Experimental evaluation reveals that a higher impact of latent space regression leads to improved results. This implies that the model, when trained with non-occluded images, produces a latent space that is both pertinent and well-suited for effective clustering.

%This suggests that training the model with non-occluded images results in a latent space that is relevant and well-suited for clustering purposes.

\subsection{Unsupervised Deep Clustering} 

\begin{figure*}[ht]
 \centering
 \includegraphics[width=1\textwidth]{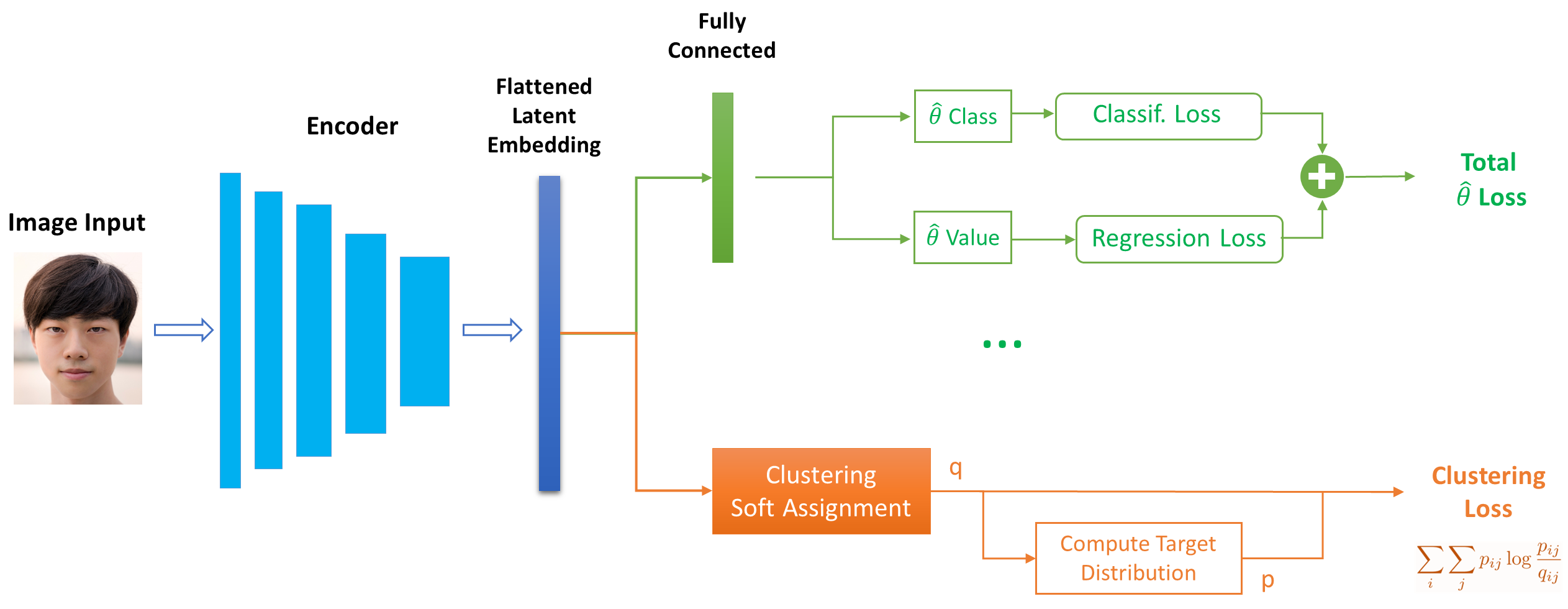}
% \vspace{-3mm}
 \caption{Network structure for LEC-HPE. The architecture includes a branch for clustering of feature space embeddings and one multi-loss branch for each predicted Euler angle ($\hat{\theta}\in \{yaw, pitch, roll\}$), to ensure continuous feature learning and avoid distortion of the latent space.}
 \label{fig: LEC-HPE}
%\vspace{-5mm}
\end{figure*}

Unsupervised clustering analysis has become quite relevant learning strategy in several fields, {\em e.g.} data science, machine learning, and computer vision, to quote a few. Within the scope of our application, it focuses on grouping image data without ground truth labels and based on some feature similarity metric. On the one hand, simpler methods like k-means~\cite{kmeans} are effective for clustering, however, struggle to learn representations and cluster data when the input feature space dimension is high. Thus, transforming the data to a lower dimensional space is a viable solution. On the other hand deep neural networks have become a fundamental tool to perform feature dimension reduction. Thus, performing data clustering in low dimensional subspaces  seems a natural approach to be adopted. One way to achieve this is to use {\em deep clustering}~\cite {ren2022deep}. Deep clustering, however, remains a relatively recent field of study, but some works are available in the field. For instance, the unsupervised Deep Embedding Clustering (DEC) method~\cite{dec} uses neural networks to achieve accurate clustering. It learns a mapping from the input data to a relevant lower dimensional feature space using an auto-encoder with an image reconstruction loss and iteratively optimizes a clustering objective. The authors of \cite{idec} argue that relying solely on the clustering loss may not guarantee the preservation of local structure, potentially resulting in the corruption of the feature space. They propose incorporating both the clustering and the reconstruction losses of the auto-encoder. This approach aims to optimize cluster assignment while concurrently updating features to preserve local structure for clustering. The work of \cite{cdec} employs augmented data and integrates the clustering loss with an instance-wise contrastive loss. This combination aims to maximize the similarity of positive pairs while penalizing negative ones. Additionally, an anchor loss is introduced to maximize agreement between raw samples and augmented cluster assignments.

%%%%%%%%%%%%%%%%%%%%%%%%%%%%%%%%%%%%%%%%%%%%%%%%%%%%%%%%%%%%%%%%%%%%%%%%%%%%%%%%
\section{Methodology}
\label{section: methodology}

The proposed methodology aims to address the difficulties mentioned in Sec.~\ref{section:Introduction}. Specifically, we aim to use few ground truth latent embeddings, say $K$, to obtain the representative of $N$ images, having $K<<N$. To accomplish this, we use unsupervised latent embedding clustering motivated by the ideas of~\cite{dec, idec, cdec}. We integrate the clustering objective with additional functions to ensure the preservation of accurate feature representation and prevent corruption of the latent feature space. In our case, this involves the use of multi-loss functions for each estimation angle. Also, as it will be seen in the experimental evaluation, we address the limitation (2), that is, to relax about the ground truth constraint requirements when augmenting the training data. 
We refer to our work as \textit{Latent Embedding Clustering for Head Pose Estimation} (LEC-HPE).

\subsection{Latent Embedding Clustering for Head Pose Estimation}

The overall architecture for the LEC-HPE model is illustrated in Fig.~\ref{fig: LEC-HPE}. This structure includes a backbone encoder and a total of four separate branches subject to optimization. Three branches are used for the prediction of each  Euler angle $\hat{\theta}\in \{yaw, pitch, roll\}$, and include fully-connected layers that output logits to be processed within a multi-loss framework for classification and fine-grained estimation. The remaining branch is responsible for clustering and fine-tuning the latent space. In the following subsections, we will describe the proposal in more detail.  

We aim to align the latent space encoding for occluded images closely with that of non-occluded images. In order to avoid requiring unique latent embedding labels for each training image, which can be computationally expensive, we perform clustering of the latent space.
Formally, let us consider a dataset $X$ with $N$ image samples $x_i$ {\em i.e.}, $\{x_i\in\ X\}_{i=1}^N$. The images in $X$ are the input to the encoder which transforms the data with a non-linear mapping $f_\Omega : X\rightarrow L$, where $\Omega$ are the learning parameters of the network and $L$ the latent feature space.  
In the $L$ space, the embedded points of semantically similar data samples are closer together. 
The proposed clustering algorithm acts on the latent feature space $L$, by learning a set of $K$ clusters centers, {\em i.e.}, $\{ c_j \in L\}_{j=1}^K$. Thus, we approximate each embedded point $l_i\in L$ to a cluster centroid $c_j$ which can be learned and refined in an unsupervised manner. The number of clusters ($K$) will be much smaller than the size of the training dataset $X$, avoiding a high computation cost. That is,  our strategy is not dependent on occluded/non-occluded image pairs and allows for the expansion of the training dataset without the need for latent labels. Thus, relaxing on the constrain data augmentation as observed, for instance, in~\cite{latent}), where each each occluded training  image requires the correspondent non-occluded embedded ground truth.  
\begin{figure*}
  \centering
  \begin{subfigure}[b]{0.49\textwidth}
    \includegraphics[width=\textwidth]{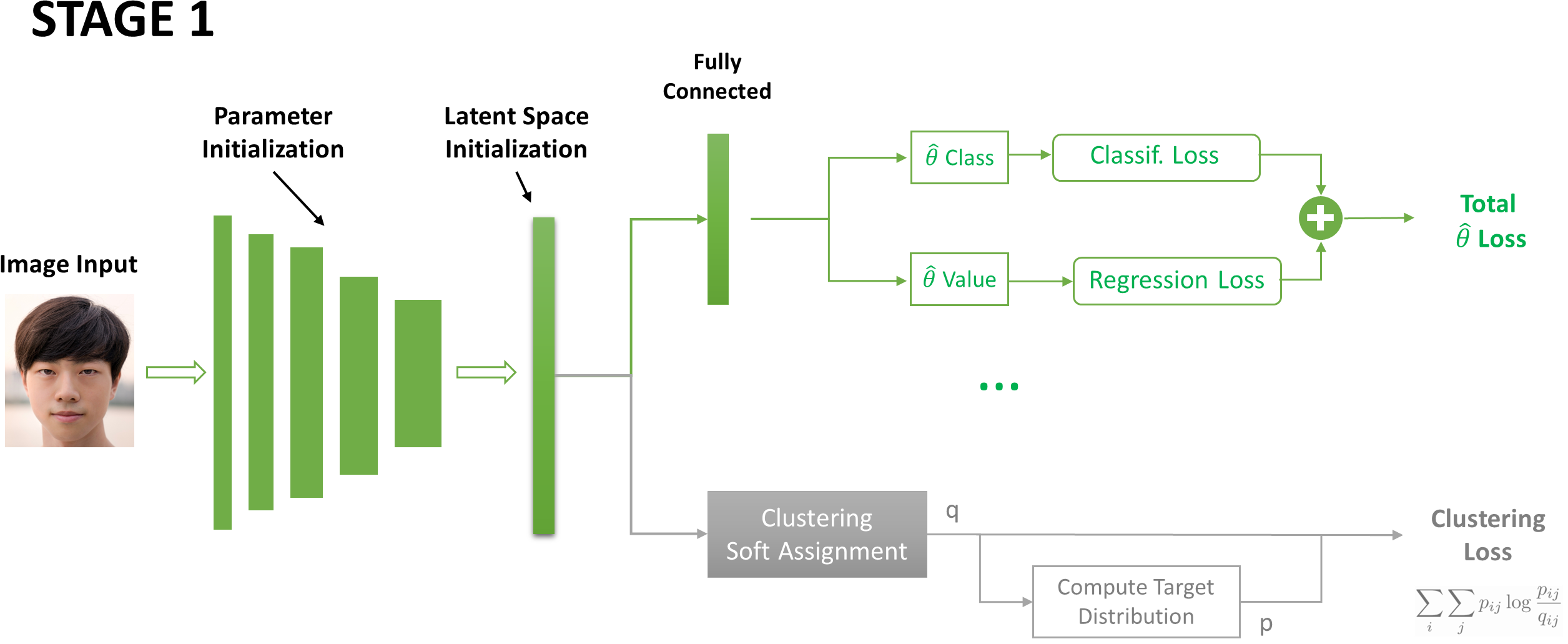}
    \caption{Stage 1: Parameter and latent space initialization.}
    \label{subfig:stage1}
  \end{subfigure}
  %\hfill
  \begin{subfigure}[b]{0.49\textwidth}
    \includegraphics[width=\textwidth]{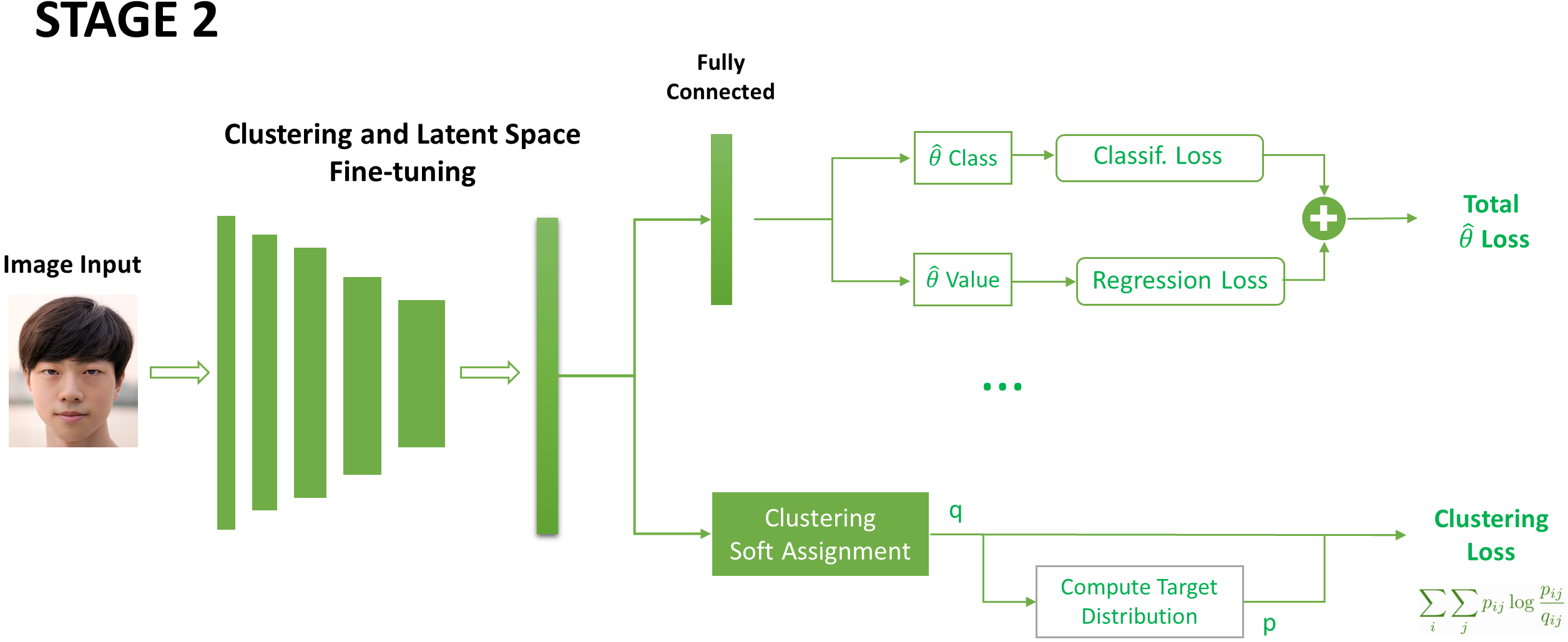}
    \caption{Stage 2: Clustering and latent space finetuning.}
    \label{subfig:stage2}
  \end{subfigure}
  \caption{Two-stage training: Green highlights indicate the model components involved in each stage, while grey highlights represent parts not included in the respective stages.}
  \label{fig:twostages}
%\vspace{-5mm}
\end{figure*}
 
\subsection{Two-stage Training}
\label{subsection: two-stage}

Our training strategy is two-fold, as illustrated in Fig.~\ref{fig:twostages}. {\bf Stage 1}: in the first stage, we perform parameter and feature space initialization {by optimizing the model for bin prediction using a classification loss and fine-grained Euler angle estimation using a regression loss; {\bf Stage 2}:  in the second stage, we add a clustering term. While in~\cite{dec} the authors drop the loss function used for parameter/feature space initialization and replace it with the clustering loss, we maintain the multi-loss functions involved in the first stage. %(1) In a first stage, we perform parameter and feature space initialization by optimizing the model with multi-loss objective functions for each fine-grained prediction of each of the Euler angles; (2) In the second stage, we add a clustering objective function. While in~\cite{dec} the authors drop the optimization objective for parameter/feature space initialization and apply only a clustering objective, we maintain the entire architecture involved in the fine-tuning process.
Similar to previous studies~\cite{cdec,idec}, we also confirm that relying solely on the clustering loss in the second stage results in the degradation of the latent feature space (see results in Sec.~\ref{sec:Impact}). The purpose of this stage is to perform clustering and fine-tune the latent space while ensuring that it preserves its relevance for the head pose estimation task.

\subsection{Fine-grained Losses for Feature Learning and Preservation}

Within the proposed architecture illustrated in Fig.~\ref{fig: LEC-HPE}, there are three branches related to the Euler angle pose estimation. The top (green) branch is dedicated to a specific Euler angle $\hat{\theta}\in \{yaw, pitch, roll\}$ and incorporates a fully-connected layer that extends the backbone network. This layer outputs a vector of raw predictions (logits) for classification purposes.} A Softmax activation function converts the logits to a vector of normalized probabilities with one value for each possible class: 

\begin{equation}
S(y_i) = \frac{e^{y_i}}{\sum_{j = 1}^{n}e^{y_j}}
\label{eq:softmax}
\end{equation}
where $y_i$ is the logit for class $i\in \{1,...,n\}$. Each class corresponds to a classification bin that covers a pose range in degrees, {\em e.g.} [0º, 3º]. From here, the output of the Softmax activation function is processed in two separate branches: one for classification and the other for regression. 
\par Regarding the classification branch, we apply a cross-entropy loss which takes the normalized probabilities and measures the distance from the ground truth values:

\begin{equation}
L_{class} = - \sum_{i}^{C}{t}_i \log\left ( S(y_i)  \right ) 
\label{eq:crossentropy}
\end{equation}
where $t_i$ is the truth value (0 or 1), C is the number of classes and $S(y_i)$ is the Softmax probability for the $i^{th}$ class. 
\par For the regression branch, we use the Softmax probabilities to compute the predicted angle in degrees from the expected value, as defined in \cite{whenet}:

\begin{equation}
\hat{\theta}_{Euler} = r \sum_{i = 1}^{C} S(y_i) \left ( i-\frac{1+C}{2} \right )
\label{eq:expectedvalue}
\end{equation}
where $\hat{\theta}_{Euler}$ is the predicted Euler angle in degrees and $r$ is the angle width of each bin, ({\em e.g.} if the bin is [0º, 3º], then $r=3$). Afterwards, we apply a mean squared error loss (MSE) as the regression objective between the predicted and ground truth Euler angle:

%$\frac{1+N}{2}$ is a constant calculated as the midpoint of bin indices. The term $i-\frac{1+N}{2}$ determines the displacement of the bin index from the center and ensures that when we multiply it by the softmax probability we are effectively assigning weights to each bin based on its position relative to the center of the range. It allows the mapping of bin indices to angles and calculation of the weighted sum to obtain the expected value of the distribution.}

\begin{equation}
L_{reg}  = \frac{1}{N}\sum_{i = 1}^{N} \left \| \theta _{i_{gt}} -  \hat{\theta}_{i_{Euler}}\right \|_{2}^{2}     
\label{eq:lossreg}
\end{equation}
where $N$ is the number of predictions, $\theta_{i_{gt}}$ is the ground truth and $\hat{\theta}_{i_{Euler}}$ the predicted Euler angle.
\par The classification and regression losses are then combined, using a regularization coefficient, to yield the final total loss for each Euler angle ($L_{yaw},L_{pitch},L_{roll}$):

\begin{equation}
\centering
\begin{split}
L_{\hat{\theta}} = L_{class\_\hat{\theta}}^{}(\hat{\theta}_{pred},{\theta}_{gt}) + \alpha \, L_{reg\_\hat{\theta}}^{}(\hat{\theta}_{pred},{\theta}_{gt})
%L_{pitch} = L_{class\_p}^{}(y_{pred},y_{gt}) + \alpha \, L_{reg\_p}^{}(y_{pred},y_{gt}) \\
%L_{roll} = L_{class\_r}^{}(y_{pred},y_{gt}) + \alpha \, L_{reg\_r}^{}(y_{pred},y_{gt}) \\
\end{split}
\label{eq:losses_angle}
\end{equation}
where $L_{class\_\hat{\theta}}$ and  $L_{reg\_\hat{\theta}}$ are classification and regression losses for $\hat{\theta} \in \{yaw, pitch, roll\}$, $\hat{\theta}_{pred}$ and ${\theta}_{gt}$ are predicted and ground truth values, respectively and $\alpha>0$ is the regularization coefficient that manages the trade-off between the two terms. The regression loss is used to achieve fine-grained estimations, while the classification loss helps the model to predict the vicinity of 
the pose. These Euler objective functions allow the model to learn the best non-linear mapping between the data space and the latent space for the task of head pose estimation during the first stage of training, while aiding the model to preserve an adequate mapping during the second stage (see results in Sec.~\ref{sec:Impact}).

\subsection{Unsupervised Latent Clustering}

The remaining branch of the architecture defined in Fig.~\ref{fig: LEC-HPE} concerns the clustering of the latent feature space. Once {\bf Stage 1} (Fig.~\ref{subfig:stage1}) of the training procedure is completed, the encoder has learned an initial non-linear mapping $f_\Omega : X\rightarrow L$, which transforms the input image data space $X$ to a relevant representation for head pose estimation in the latent space $L$. Therefore, we can move to {\bf Stage 2} (Fig.~\ref{subfig:stage2}) and perform clustering during training to fine-tune the feature space. As in \cite{dec}, we perform k-means~\cite{kmeans} in the latent feature space mapped from the entire training input data $Z = f_\Omega\left ( X\right )$, to obtain the initial cluster centers $\{c_i\}_{i=1}^{K}$, where $K$ is the number of clusters.
\par With initialized cluster centroids, we measure the pairwise similarity $q_{ij}$ between a latent embedded point $l_i$ and a cluster center $c_j$ according to \cite{dec} and also used in the t-Distributed Stochastic Neighbor Embedding (t-SNE) \cite{tsne} technique: 

\begin{equation}
q_{ij} = \frac{(1+\left \| l_{i}-c_{j} \right \|^{2})^{-1}}{\sum_{k\neq l}^{}(1+\left \| l_{k}-c_{l} \right \|^{2})^{-1}}
\label{eq:soft_assignment}
\end{equation}

\par As described in \cite{dec}, this quantified similarity can be interpreted as the probability of assigning the $i^{th}$ sample to the $j^{th}$ cluster center and therefore considered a soft/probabilistic assignment. From this soft assignment $q_{ij}$ we can compute the target distribution $p_{ij}$:

\begin{equation}
p_{ij} = \frac{q_{ij}^2/\sum_{i}^{}q_{ij}}{\sum_{j'}(q_{ij'}^2 /\sum_{i}q_{ij'})}
\label{eq:target distribution}
\end{equation}

With $q_{ij}$ and $p_{ij}$, we can optimize cluster centers and fine-tune the encoder parameters by minimizing a Kullback-Leibler (KL) divergence objective between the target distribution $P$ and soft assignment $Q$:

\begin{equation}
L_{clustering} = {\rm KL}(P\parallel Q) = \sum_{i}^{}\sum_{j}^{}p_{ij} \log \frac{p_{ij}}{q_{ij}}
\label{eq:target distribution}
\end{equation}

\subsection{Overall Loss}

The overall loss is the sum of all described objective functions with regularization coefficient $\beta$ to adjust the impact of the clustering term in the optimization of the model:
%\vspace{-1mm}
\begin{equation}
L_{total} = L_{yaw}+ L_{pitch}+ L_{roll}+ \beta L_{clustering}
\label{eq:overall loss}
\end{equation}
where $\beta = 0$ during the first training stage, and $\beta > 0$ during the second training stage. 

%-------------------------------------------------------------------------
\section{Experimental Evaluation}
This section describes carefully the extensive experimental evaluation  involving  the use of several benchmark HPE datasets.

\label{section: experimental evaluation}

\begin{table*}[ht]
    %\begin{center}
    \centering
    \resizebox{1\linewidth}{!}{
    \begin{tabular}{c|cccc|cccc|c}
    BIWI\cite{biwi}          & \multicolumn{4}{c|}{Non-occluded Images} & \multicolumn{4}{c|}{Occluded Images} & Combined \\ \hline
                  & Yaw      & Pitch    & Roll     & MAE     & Yaw     & Pitch   & Roll    & MAE    & MAE      \\ \hline
    FSA-Net \cite{FSA_NET}       & 5.420    & 5.568    & 4.515   & 5.168   & 10.987  & 9.848   & 7.846   & 9.560  & 7.364    \\
    6DRepNet \cite{6drepnet}      & 4.238  & 4.580   & 3.337  & 4.051  & 7.883   & 14.983  & 9.665   & 10.844 & 7.252    \\
    DAD-3D \cite{dad3d}        & \bf 3.668 &  3.855    & 3.537    & \bf 3.687   & \bf 5.532   & 7.924   & 7.478   & 6.978  & 5.328   \\
    Lightweight \cite{lightweight}   & 5.135  & 3.561   &  4.106 &  4.267 & 10.784  & 9.654   & 10.309  & 10.249 & 7.292    \\
    Hopenet \cite{hopenet}       & 4.375    & \bf 3.559    &  \bf 3.348   & 3.761   & 6.725   & 8.616   & 7.338  & 7.560  & 5.661  \\
    LSR \cite{latent}           &  \textcolor{gray}{\textit{4.291}}    & \textcolor{gray}{\textit{3.086}}    & \textcolor{gray}{\textit{3.179}}    & \textcolor{gray}{\textit{3.519}}   & \textcolor{gray}{\textit{5.429}}   & \textcolor{gray}{\textit{4.823}}   & \textcolor{gray}{\textit{3.467}}   & \textcolor{gray}{\textit{4.573}}  & \textcolor{gray}{\textit{4.046}}    \\
    LEC-HPE (ours)     & 4.602    & 3.897    & 3.759    & 4.086   & 6.153   & \bf 6.576   & \bf 4.605   &  \bf 5.778 & \bf  4.932  
    %LEC-HPE (ours)     & 4.602}    & 3.897}    & 3.759}    & 4.086}   & 6.153}   & \bf 6.576}   & \bf 4.605}   &  \bf 5.778}  & \bf  4.932}   
    \end{tabular}
    }
    %\end{center}
    \caption{Head pose estimation MAE (º) results with BIWI. LSR method values are in grey italics since they constitute upper-bound target performance. Bold values are best results apart from target performance.}
    \label{table biwi}

\vspace{3mm}
\end{table*}

\begin{table*}[ht]
\centering
\resizebox{1\linewidth}{!}{
\begin{tabular}{c|cccc|cccc|c}
AFLW2000 \cite{aflw}      & \multicolumn{4}{c|}{Non-occluded Images} & \multicolumn{4}{c|}{Occluded Images} & Combined \\ \hline
              & Yaw      & Pitch    & Roll     & MAE     & Yaw     & Pitch   & Roll    & MAE    & MAE      \\ \hline
FSA-Net \cite{FSA_NET}       & 5.109    & 6.462    & 3.356    & 5.642   & 13.664  & 10.880  & 10.067  & 11.537 & 8.590    \\
6DRepNet \cite{6drepnet}      & 3.230    & 4.658    & 3.091    & 3.660   & 8.904   & 9.799   & 7.408   & 8.704  & 6.182          \\
DAD-3D \cite{dad3d}        & \bf 3.134    & \bf 4.630    & \bf 3.090    & \bf 3.618   & 7.084   & 14.953  & 13.680  & 11.906 & 7.762         \\
Lightweight \cite{lightweight}   & 4.267    &  5.015   & 3.722    & 4.335  & 10.771  & 9.278   & 8.335   & 9.461  & 6.898         \\
Hopenet \cite{hopenet}       & 4.965    & 5.250    & 3.956   & 4.724   & 12.438  & 10.277  & 8.586   & 10.434 & 7.579    \\
LSR~\cite{latent}          & \textcolor{gray}{\textit{3.813}}  & \textcolor{gray}{\textit{5.420}}    & \textcolor{gray}{\textit{4.003}}    & \textcolor{gray}{\textit{4.412}}   & \textcolor{gray}{\textit{4.741}}   & \textcolor{gray}{\textit{6.254}}   & \textcolor{gray}{\textit{4.765}}   & \textcolor{gray}{\textit{5.253}}  & \textcolor{gray}{\textit{4.833}}    \\
LEC-HPE (ours)   & 3.843    & 5.177   & 4.038    &  4.353   & \bf 5.211   & \bf 6.481   & \bf 5.064   & \bf 5.589  & \bf 4.971     
\end{tabular} 
}
\caption{Head pose estimation MAE (º) results with AFLW2000. LSR method values are in grey italics since they constitute upper-bound target performance. Bold values are best results apart from target performance.}
\label{table aflw}
\vspace{3mm}
\end{table*}

\begin{table}[h]
\centering
\resizebox{1\linewidth}{!}{
\begin{tabular}{c|cccc} 
Pandora \cite{pandora}       & \multicolumn{4}{c}{Occluded Images} \\ \hline
              & Yaw      & Pitch  & Roll   & MAE    \\ \hline
FSA-Net \cite{FSA_NET}       & 11.736   & 8.607  & \bf 6.691  & 9.011  \\
6DRepNet \cite{6drepnet}     & 10.896   & 6.897  & 7.500  & 8.431  \\
DAD-3D \cite{dad3d}       & 9.348    & 7.437  & 8.474  & 8.420  \\
Lightweight \cite{lightweight}   & 10.133   & 8.690  & 7.064  & 8.629  \\
Hopenet \cite{hopenet}      & 10.442  & 7.239  & 6.925  & 8.202  \\
LSR \cite{latent}       & \textcolor{gray}{\textit{9.096}}   & \textcolor{gray}{\textit{5.657}}  & \textcolor{gray}{\textit{6.215}}  & \textcolor{gray}{\textit{6.989}}  \\
LEC-HPE (ours)      & \bf 9.071    & \bf 5.885  &  6.918 & \bf 7.291  
\end{tabular}
}
\caption{Head pose estimation MAE (º) results with Pandora. LSR method values are in grey italics since they constitute upper-bound target performance. Bold values are best results apart from target performance.}
\label{table pandora}
%\vspace{-5mm}
\end{table}

\subsection{Datasets}
\par We utilize four distinct benchmark in-the-wild head pose datasets for training and testing purposes: 300W-LP \cite{300wlp}, BIWI \cite{biwi}, AFLW2000 \cite{aflw} and Pandora \cite{pandora}. 
\vspace{2mm}
\begin{itemize}

\item The \textbf{300W-LP dataset} is a synthetic dataset used for training and comprises an extensive collection of 61225 face samples. It spans a wide range of individuals, illumination conditions, and poses, which makes it ideal for training head pose estimation models \cite{whenet, latent}. 
\vspace{2mm}
\item The \textbf{BIWI dataset} is used for testing and offers over 15000 images that capture head poses from 20 individuals along wide ranges, mainly for yaw and pitch.  Widely recognized as a benchmark dataset for head pose estimation challenges \cite{hopenet, quatnet,whenet}, it provides depth and RGB images (640x480 pixels) and ground truth pose annotations for each image. 
\vspace{2mm}
\item The \textbf{AFLW2000 dataset} is used for testing and consists of 2000 in-the-wild images with diverse head poses and strong variations in lighting and background conditions. It contains ground truth annotations of the Euler angles of the head poses. 
\vspace{2mm}
\item For all the aforementioned datasets, and similarly to ~\cite{latent}, we generate synthetic occlusions to obtain the occluded versions of the 300W-LP, BIWI, and AFLW2000 datasets.
\vspace{2mm}
\item The \textbf{Pandora dataset} \cite{pandora} is used for testing and includes real-life occlusions. It simulates driving poses from the perspective of a camera situated inside a dashboard. It contains over 250000 RGB (1920x1080 pixels) and depth images (512x424) with corresponding annotations, capturing head poses in a wide range of the Euler angles. For our specific tests, we use 9619 occluded head images, featuring actors wearing various common occlusion garments such as sunglasses, scarves, caps, and masks.
\end{itemize}

\subsection{Implementation Details}
\label{implementation details}

In the network structure illustrated in Fig.~\ref{fig: LEC-HPE}, we use ResNet-50 \cite{resnet50} as the backbone encoder. This 50-layer convolutional neural network has proven to be effective and suitable for head pose estimation within deep learning frameworks~\cite{hopenet,latent}. Specifically, we use a ResNet-50 pre-trained on ImageNet~\cite{imagenet}. We follow the training procedure described in Section~\ref{subsection: two-stage} and illustrated in Fig.~\ref{fig:twostages}. In {\bf Stage 1}, we use the original non-occluded 300W-LP dataset to initialize the encoder parameters and the latent feature space. In {\bf Stage 2}, we use the synthetically occluded version of the same dataset. Both training stages are trained for 25 epochs, using Adam optimization with a scheduled learning rate initialized at $10^{-5}$, $\epsilon = 10^{-8}$, ${\beta}_1=0.9$ and ${\beta}_2=0.999$. The regularization coefficient $\alpha$ for the regression components (see (\ref{eq:losses_angle})) is set to 1. We trained our model with initialization for $K=10$ clusters, which led to optimal results
(see Sec.~\ref{sec:HPE-for-three--datasets}). All training and testing input images are pre-processed using a face detector to crop the face region.

Opposing to~\cite{latent}, we no longer need to use the latent ground truth of the images with which we train the {\bf Stage 1}. Thus, we no longer depend on image pairs ({\em i.e.}, non-occluded and respective occluded replicas) and the correspondent ground truth {\em constraint} requirements when augmenting the training data (see limitation {\bf (2)} in Sec.\ref{section:Introduction}). \par The images are also resized to the input dimension of the ResNet-50 backbone ({\em i.e.} 224x224x3) and normalized using the mean and standard deviation of ImageNet for all color channels. We train our model with 80-20 training/validation splits and a training batch size of 128. The fully-connected layers predict 66 different classification bins in the range $\pm 99$º. We augment the training dataset by randomly flipping, blurring, down-sampling, and up-sampling images.

\subsection{Head Pose Estimation Results on the BIWI, AFLW2000 and Pandora Datasets}
\label{sec:HPE-for-three--datasets}

We assess the estimation error of our methodology and compare it to the state-of-the-art methods in both the original and synthetic occluded variants of the BIWI and AFLW2000 datasets. Additionally, we evaluate the performance on the Pandora dataset, which includes real-life occlusions. Regarding the AFLW2000 datasets, 31 images were excluded as they fell outside the range accommodated by our bin classification. The head pose estimation mean squared error (MAE) results are displayed in Tables~\ref{table biwi},~\ref{table aflw} and~\ref{table pandora}, for BIWI, AFLW2000, and Pandora, respectively. 

Notice that in Tables~\ref{table biwi},~\ref{table aflw} and~\ref{table pandora} we include the method LSR in ~\cite{latent}, as this method uses a ground truth for each image, that is, having $K=N$. For instance, in the case of 300W-LP training dataset, this involves $K=61225$ embedding ground truth samples. Our approach only requires $K=10$, representing cluster centroids. This means that the results in~\cite{latent} are derived by accessing the entire latent information, whereas our method only has access to a much smaller number of cluster centroids in that latent space. Thus, the results of the method in ~\cite{latent} constitute our upper bound performance since we do not anticipate our method to outperform these results. Our goal is to closely approach their performance while significantly reducing the required ground truth data. 

To obtain the optimal number of $K$ centroids, we applied the elbow method~\cite{elbow} with the initialized latent feature space and tested HPE in models with $K\in\{5,10,15,20,30,40\}$. Both the elbow method and the tests indicated $K=10$ as the model order that provides the best performance.

Overall, our method delivers competitive performance compared to \cite{latent} and surpasses other SoTA methodologies in occluded scenarios by a significant margin. 

\par When compared to the upper-bound method (LSR), the occluded average MAE was approximately only 0.3º worse in AFLW2000 and Pandora. Furthermore, the non-occluded average error was identical to LSR and even surpassed it in AFLW2000. This led to a combined average MAE which is also competitive with LSR results across all datasets. The combined average MAE for AFLW2000 is only 2.9\% worse than that of LSR. Regarding other state-of-the-art methods, the LEC-HPE estimation results in occluded images improve results by a substantial margin of 36\%, 17\%, and 11\%, when compared to the best-performing method (apart from upper-bound LSR) in AFLW2000, BIWI, and Pandora, respectively. Regarding the combined MAE, our model surpasses the best-performing method (apart from LSR) by 19.6\% in AFLW2000, 11.1\% in Pandora, and 7\% in BIWI.

% BIWI OCCLUDED 8.9% 0.474º 17% COMBINED 5.6% 0.263º 7%
% AFLW OCCLUDED 6% 35.8% 0.336º COMBINED 2.9% 0.138º 19.6%
% PANDORA OCCLUDED 4.3% 11.1% 0.302º COMBINED 4.3% 0.302º 11.1%

\subsection{Study on the Impact of Clustering}\label{sec:Impact}

\begin{table}[h]
\centering
\resizebox{0.95\linewidth}{!}{
\begin{tabular}{c|ccc}
%          & \multicolumn{3}{c}{BIWI \cite{biwi}}      \\ \hline
BIWI \cite{biwi}         & Non-occluded & Occluded & MAE \\ \hline
$\beta=0$    &    4.653     &   6.434       & 5.544    \\
$\beta=10$   &    4.509     &   6.096       &  5.303   \\
$\beta=100$  &    \textbf{4.086}     &   \textbf{5.778}  &  \textbf{4.932}   \\
$\beta=1000$ &    5.035     &   6.676       &  5.856  
\end{tabular}
}
\caption{LEC-HPE MAE (º) with different $\beta$ values in the BIWI dataset.}
\label{beta_biwi}
%\vspace{-4mm}
\end{table}

\begin{table}[h]
\centering
\resizebox{0.95\linewidth}{!}{
\begin{tabular}{c|ccc}
%AFLW2000 \cite{aflw}          & \multicolumn{3}{c}{}      \\ 
AFLW2000 \cite{aflw}           & Non-occluded & Occluded & MAE \\ \hline
$\beta=0$    &  5.069    &   6.243       & 5.656     \\
$\beta=10$   &  4.987    &   6.095       & 5.541    \\
$\beta=100$  &  \textbf{4.353}    &   \textbf{5.589}       &   \textbf{4.971}    \\
$\beta=1000$ &  5.610    &   6.593       & 6.102   
\end{tabular}
}
\caption{LEC-HPE MAE (º) with different $\beta$ values in the AFLW2000 dataset.}
\label{beta_aflw}
%\vspace{-4mm}
\end{table}

%\begin{table}[h]
%\centering
%\resizebox{0.95\linewidth}{!}{
%\begin{tabular}{c|cccc}
%          & \multicolumn{4}{c}{Pandora \cite{pandora}}      \\ \hline
%          &  $\beta=0$  & $\beta=10$ & $\beta=100$ & $\beta=1000$ \\ \hline
%Occluded &  7.938    &   7.787       & \textbf{7.291} & 8.169   
%\end{tabular}
%}
%\caption{LEC-HPE MAE (º) with different $\beta$ values in the Pandora dataset.}
%\label{beta_pandora}
%\end{table}

\begin{table}[h]
\centering
\resizebox{0.55\linewidth}{!}{
\begin{tabular}{c|ccc}
%          & \multicolumn{1}{c}{Pandora \cite{pandora}}      \\ \hline
Pandora \cite{pandora}          & Occluded  \\ \hline
$\beta=0$    &  7.938     \\
$\beta=10$   &  7.787   \\
$\beta=100$  &  \textbf{7.291}      \\
$\beta=1000$ &  8.169   
\end{tabular}
}
\caption{LEC-HPE MAE (º) with different $\beta$ values in the Pandora dataset.}
\label{beta_pandora}
%\vspace{-4mm}
\end{table}

We carried out an ablation study on the impact of the clustering term by varying the $\beta$ regularization coefficient in the overall training loss implemented in {\bf Stage 2} (see (\ref{eq:overall loss})). We trained LEC-HPE with four different values for this coefficient, $\beta= \{ 0,10,100,1000\}$. The MAE estimation results are listed in Tables~\ref{beta_biwi}, \ref{beta_aflw} and \ref{beta_pandora}. 
\par When $\beta=0$, the second training stage does not include the clustering loss and is identical to the first training stage. This means that we only use the top branch in Fig.~\ref{fig: LEC-HPE}. 

The results across all datasets reveal that including the clustering term ($\beta=10$ and $\beta=100$) in the second stage reduces the estimation error for occluded and non-occluded images. In particular, when $\beta=100$ is employed, optimal outcomes are observed, resulting in an error reduction ranging from $8\%$ to $12\%$ compared to when the clustering term is not utilized ($\beta=0$). Estimation errors escalate with $\beta=1000$, indicating that an excessively strong influence of the clustering term on the overall loss leads to the deterioration of the properly initialized latent feature space. This confirms the need to maintain the fine-grained Euler angle losses in the second training stage to avoid the model deviating from the best performance for the fine-grained head pose estimation task.

%\subsection{Ablation Study of Backbone Encoder}

%\begin{table}[h]
%\begin{tabular}{c|ccc}
%          & \multicolumn{3}{c}{BIWI \cite{biwi}}      \\ \hline
%          & Non-occluded & Occluded & MAE \\ \hline
%EFN B0    &              &          &     \\
%EFN B1    &              &          &     \\
%EFN B2    &              &          &     \\
%EFN B3    &              &          &    \\
%EFN B3    &              &          &    \\
%RESNET50  &              &          &    \\

%\end{tabular}
%\caption{LEC-HPE with different backbone encoders with BIWI dataset.}
%\end{table}

%\begin{table}[h]
%\begin{tabular}{c|ccc}
%          & \multicolumn{3}{c}{AFLW2000 \cite{aflw}}      \\ \hline
%          & Non-occluded & Occluded & MAE \\ \hline
%EFN B0    &              &          &     \\
%EFN B1    &              &          &     \\
%EFN B2    &              &          &     \\
%EFN B3    &              &          &    \\
%EFN B3    &              &          &    \\
%RESNET50  &              &          &    \\

%\end{tabular}
%\caption{LEC-HPE with different backbone encoders with AFLW dataset.}
%\end{table}

\section{Conclusion}
\label{section: conclusion}
%\vspace{4mm}

In this paper, we present an efficient methodology to address the challenge of occlusion in head pose estimation, a significant hurdle in this sub-field of computer vision. From the experimental evaluation, the proposed framework improves occlusion robustness by combining unsupervised latent embedding clustering in the latent feature space with a fine-grained Euler angle multi-loss scheme. 

The main idea behind LEC-HPE is to improve feature representation for pose estimation while refining the latent embedding space through clustering, eliminating the need for labeled embedding data for each training image. This approach offers a more efficient alternative compared to some of the most recent occlusion-focused state-of-the-art work, without requiting a constrained expansion of the training dataset.
\par We demonstrate experimentally that we can achieve similar results without the need to have ground truth for each latent embedding label. Our results also surpass the state of the art for standard head pose estimation by a significant margin in occluded images. We perform an ablation study to quantitatively evaluate the impact of the clustering term and verify that using this term improves the estimation results. We also confirm the need  to include the fine-grained Euler angles scheme to avoid latent space corruption. 
\par Although not fully addressed in this paper, further work will include a process for automatic selection of the optimal parsimonious number of cluster centroids in the low dimensional latent space. Also, we plan to test this methodology with smaller and more efficient backbone encoders for low-power applications. We also intend to evaluate the use of clustering losses for the classification component in the multi-loss scheme.

\section{Acknowledgements}
This work was supported by LARSyS funding (DOI: 10.54499/LA/P/0083/2020, 10.54499/UIDP/50009/2020, and 10.54499/UIDB/50009/2020], through Fundação para a Ciência e a Tecnologia and by the SmartRetail project [PRR - C645440011-00000062], through IAPMEI - Agência para a Competitividade e Inovação.

{\small
\bibliographystyle{ieee}
\bibliography{egbib}
}

\end{document}